\documentclass[sigconf]{acmart}
\setcopyright{none}
\renewcommand\footnotetextcopyrightpermission[1]{}
\acmConference[MFMB '26]{Workshop on Multimodal and Foundation Models in Banking at ICAIF '26}{November 15, 2026}{Milan, Italy}

\usepackage{booktabs}
\usepackage{tikz}
\usetikzlibrary{arrows.meta}
\usepackage{float}
\usepackage{etoolbox}

\newcommand\aff[1]{\texorpdfstring{\textsuperscript{#1}}{}}
\newcommand\affsep{,}
\makeatletter
\def\@ACM@checkaffil{}
\newcount\@intentauthor
\patchcmd{\@typeset@author@bx}
  {\@authorfont\@currentauthors\par\@affiliationfont\@currentaffiliation}
  {%
   \ifdim\author@bx@wd>.5\textwidth
   \@authorfont{\@intentauthor=\z@
     \def\par{\advance\@intentauthor\@ne
       \ifnum\@intentauthor=4 ,\break\else\ifnum\@intentauthor>1 , \fi\fi}%
     \baselineskip=1.25\baselineskip\@currentauthors\endgraf}\par
   \vspace{.35\baselineskip}
   \@affiliationfont{\def\par{\unskip\hspace{1.5em}}%
     \leavevmode\@currentaffiliation\unskip}\par
   \vspace{.75\baselineskip}\fi}
  {}{\ClassError{acmart}{INTENT author-block patch failed}{Check the main.tex preamble.}}
\makeatother

\begin{document}

\title{INTENT: Out-of-Sample, Out-of-Time, and Cold-Start Evaluation on User Event Sequences}

\author[Mattias Kvist]{Mattias Kvist\aff{1\affsep}}
\authornote{These three authors contributed equally to this work.}
\author[Erik Lidman Hillbom]{Erik Lidman Hillbom\aff{1\affsep}}
\authornotemark[1]
\author[Edoardo de Cal]{Edoardo de Cal\aff{1\affsep}}
\authornotemark[1]
\author[Davis Siemens]{Davis Siemens\aff{1}}
\author[Adam Fredriksson]{Adam Fredriksson\aff{2}}
\author[Rickard Br\"uel-Gabrielsson]{Rickard Br\"uel-Gabrielsson\aff{2\affsep3\affsep}}
\authornote{Corresponding author: rickard@unboxai.com.}
\affiliation[obeypunctuation=true]{\aff{1}\institution{KTH AI Society}}
\affiliation[obeypunctuation=true]{\aff{2}\institution{Unbox AI}}
\affiliation[obeypunctuation=true]{\aff{3}\institution{MIT}}

\begin{abstract}
Banks now train foundation models on ordered sequences of customer events
such as transactions, app sessions, and authorizations. Deployed, such a model
predicts forward in time, often for customers and merchants absent from
training, yet each is evaluated on private data, so no two can be compared.
Public benchmarks for such sequences rarely test a period the model has not
seen; recommendation studies usually hold out each user's last events instead.
We present INTENT (INTEraction, Novelty, and Transfer evaluation for
next-action prediction), a public benchmark that scores a model in seven cells
defined by whether the user, the period, and the target item appeared in
training. An audit of 59 public dataset families retains five datasets; the
most common defect is the lack of a calendar timestamp that orders events. No
date lies within the time span of all five datasets, so each is split at its
own cutoff, chosen by one rule fixed before any model is evaluated. Summed over
the datasets, every cell holds at least 111{,}769 users. In reference results
the cell determines the best model: a transformer leads on seen items after
the cutoff but scores zero on new items, a variant that embeds item
metadata helps mainly on MicroLens, and repeating the user's last item leads on new
items in four of five datasets. This parameter-free control also shows that on
three datasets most of the score lost after the cutoff reflects changed
behavior rather than the model. The pipeline, the cutoff rule,
and the per-cell user counts are publicly available.
\end{abstract}

\maketitle

\section{Introduction}

Banks and payment networks now train foundation models on card transactions,
app sessions, and authorizations
\cite{ostroukhov2026pragma,braithwaite2025nuformer,dou2025transactiongpt,yeh2025treasure,stripe2025pfm,skalski2023purchasing},
and one backbone often serves fraud detection, credit, and personalization.
These models, also called large behavioral models
\cite{bruelgabrielsson2025behaviorgpt,gabrielsson2026lbm,gabrielsson2026scaling,bruelgabrielsson2026behaviorgptv4},
learn from chronological sequences of user events. Deployed, the backbone
predicts forward in time: a model trained on transactions up to June must
score, in July, a customer who opened an account in July at a merchant it has
never seen. Each bank reports results on proprietary data with its own
metrics, so no public benchmark compares such models.

Public data cannot reproduce a bank's ledger, but it can supply a task that
any group can rerun. Sequential recommendation offers such data at scale, but
its standard protocol is flawed: 77\% of recent sequential recommendation
papers hold out each user's final interactions \cite{gusak2025split}. Training
then contains events that occur after some test events, so a model can score
well by learning from the future \cite{ji2023leakage,sun2023freshlook}. The
choice of split alone can reorder models \cite{meng2020splitting}, and for
sequential models only a global time split evaluates a period the model has
not seen \cite{gusak2025split}. RecNextEval enforces such a timeline for single
datasets \cite{ng2026recnexteval}. Benchmarks for financial event sequences
address other tasks: EBES classifies whole sequences and splits by time only
where a dataset allows it \cite{osin2025ebes}, and the benchmark released with
the Multimodal Banking Dataset assigns clients to folds \cite{mollaev2025mbd}.
Neither separates new users, future periods, and new items.

Building such a test takes more than changing the split. A time split needs a
calendar timestamp that orders every event, which nearly a third of the public
dataset families we examined lack. Datasets that have one cover different
years, so no single cutoff date serves them all. A split can also leave a
condition, such as new users choosing new items, with too few users to
measure.

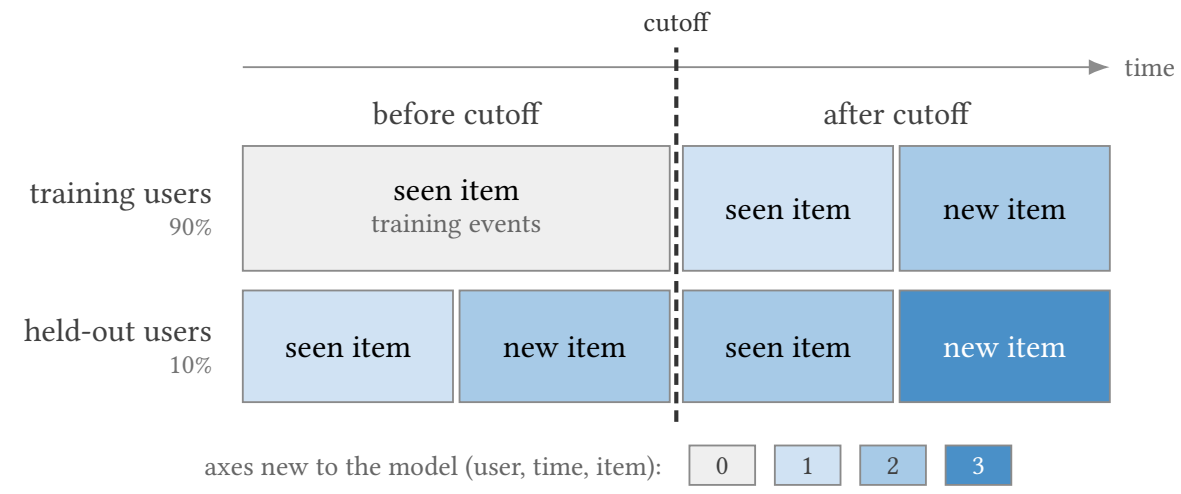
\begin{figure}[t]
\centering
\definecolor{cellnew0}{HTML}{EFEFEF}
\definecolor{cellnew1}{HTML}{D0E2F3}
\definecolor{cellnew2}{HTML}{A7CAE7}
\definecolor{cellnew3}{HTML}{4A90C8}
\definecolor{cellline}{HTML}{8A8A8A}
\resizebox{0.92\columnwidth}{!}{%
\begin{tikzpicture}[font=\footnotesize,
  cell/.style={draw=cellline, line width=0.4pt, align=center, inner sep=1pt},
  sub/.style={font=\scriptsize, text=black!60},
  key/.style={sub, anchor=west, text height=1.4ex, text depth=0.3ex},
  lab/.style={text=black!75}]
\node[cell, fill=cellnew0, minimum width=3.25cm, minimum height=0.95cm]
  at (1.625,1.475) {seen item\\[-1pt] {\scriptsize\color{black!60}training events}};
\node[cell, fill=cellnew1, minimum width=1.6cm, minimum height=0.95cm] at (4.15,1.475) {seen item};
\node[cell, fill=cellnew2, minimum width=1.6cm, minimum height=0.95cm] at (5.8,1.475) {new item};
\node[cell, fill=cellnew1, minimum width=1.6cm, minimum height=0.85cm] at (0.8,0.425) {seen item};
\node[cell, fill=cellnew2, minimum width=1.6cm, minimum height=0.85cm] at (2.45,0.425) {new item};
\node[cell, fill=cellnew2, minimum width=1.6cm, minimum height=0.85cm] at (4.15,0.425) {seen item};
\node[cell, fill=cellnew3, text=white, minimum width=1.6cm, minimum height=0.85cm] at (5.8,0.425) {new item};
\node[lab, anchor=east, align=right] at (-0.12,1.475) {training users\\[-1pt] {\scriptsize\color{black!55}90\%}};
\node[lab, anchor=east, align=right] at (-0.12,0.425) {held-out users\\[-1pt] {\scriptsize\color{black!55}10\%}};
\node[lab] at (1.625,2.2) {before cutoff};
\node[lab] at (4.975,2.2) {after cutoff};
\draw[-{Latex[length=1.6mm]}, black!45] (0,2.55) -- (6.6,2.55) node[right, sub] {time};
\draw[black!80, line width=0.9pt, dash pattern=on 2.5pt off 1.5pt] (3.3,-0.15) -- (3.3,2.7)
  node[above, sub, text=black!80] {cutoff};
\node[key, anchor=east] at (3.3,-0.48) {axes new to the model (user, time, item):};
\foreach \n/\x/\ink in {0/3.65/black!75, 1/4.3/black!75, 2/4.95/black!75, 3/5.6/white} {
  \node[cell, fill=cellnew\n, minimum width=0.5cm, minimum height=0.3cm,
    font=\scriptsize, text=\ink] at (\x,-0.48) {\n};
}
\end{tikzpicture}%
}
\caption{The seven test cells, laid out as in Figure~\ref{fig:results}. Rows
separate training users from a 10\% held-out sample, and columns separate
events before and after a per-dataset cutoff. Each row and period is divided
into seen and new target items, except the grey cell, which repeats training
events and so holds only seen items. Shading counts how many of the user,
time, and item axes are new to the model.}
\label{fig:cells}
\end{figure}

INTENT (INTEraction, Novelty, and Transfer evaluation for next-action
prediction) asks whether a model predicts the next item from ordered history
(interaction), whether that holds for new users, future periods, and new items
(novelty), and whether it holds across platforms under one protocol
(transfer). We make three contributions.
\begin{enumerate}
\item A grid of seven test cells that separates known from new users, the
training period from the future, and seen from new items
(Figure~\ref{fig:cells}), instantiated on datasets with no common time span by
a per-dataset cutoff rule (Section~\ref{sec:split}).
\item An audit of 59 public dataset families that records a reason for every
exclusion and retains five datasets (Section~\ref{sec:select},
Table~\ref{tab:families}).
\item Reference results for four models, which show that the cells reorder
them (Section~\ref{sec:baselines}).
\end{enumerate}
The code, split definitions, and per-cell user counts are available at
\url{https://github.com/kthaisociety/intent}.

\section{The evaluation grid}

INTENT scores next-item prediction. In a bank, a user is a customer, and an
item can be the merchant of the next transaction or a coarser label such as
its category. New items arise mainly at the merchant level, since merchants
open continually and category codes rarely change. A model reads a user's
ordered events and ranks the full item vocabulary of the dataset, which avoids
sampled negatives that can reverse the order of models
\cite{krichene2020sampled}. The primary metric is NDCG@10, averaged over users, which gives
more weight to correct predictions near the top of the ranking.

Three binary axes define the cells (Figure~\ref{fig:cells}). The user axis
holds 10\% of users out of training, the time axis splits events at a cutoff
date, and the item axis records whether the target item appeared in training.
The \emph{known-user, past} cell repeats training events and tests whether a
model fits what it saw; its items are all seen, so it is the one cell not
divided by item. The other six cover known users after the cutoff and new
users before and after it, each choosing a seen or a new item. These are the
conditions a deployed bank model meets: the next transaction of an existing
customer, the first months of a new customer, and a merchant that opened after
training, met by existing and new customers alike. The known-user, future cell
is the counterpart of the common per-user leave-last-out protocol, except that
no training event follows any test event. The difference between a model's
scores on seen and new items, the memorization gap, measures how much of its
score depends on memorized items. Because all five datasets share one schema,
the cells also support a transfer test in which a model trained on one dataset
is scored on another, where every target item is new
(Section~\ref{sec:discussion}).

A cell ranks models only if it holds enough distinct users. Metrics are
averaged per user and a user's events are correlated, so users are the unit of
evidence. The floor is 1{,}000 users per cell: for per-user NDCG with a
standard deviation of 0.2 to 0.3, 1{,}000 users give an approximate 95\%
interval with half-width 0.012 to 0.019.

\section{Selecting data}\label{sec:select}

Because the benchmark predicts only the next item, any dataset with users,
items, and ordered events is a candidate. We examined 101 public releases,
grouped into 59 families in which variants of one source, such as the
categories of Amazon Reviews, count once. A dataset enters only if it meets
five criteria. (i)~\emph{Stable entities}: the sequence key is a user or
account with a documented persistence assumption. (ii)~\emph{Calendar
timestamps}: every event carries a calendar time fine enough to order one
user's events; an order index or a position within a session cannot place two
users on one timeline. (iii)~\emph{Sequence population}: at least 25{,}000
users have three or more events (Figure~\ref{fig:depth}), so a 10\% sample
gives a standard error near one percentage point on a binary proportion.
(iv)~\emph{Action mapping}: each action maps to a named source field.
(v)~\emph{Access}: another researcher can obtain the same release under
documented terms. We report dataset statistics recorded during our selection audit, using a
consistent definition of sequence population: users with at least three
events in the examined release.

\begin{figure*}[t]
\centering
\includegraphics[width=0.8\textwidth]{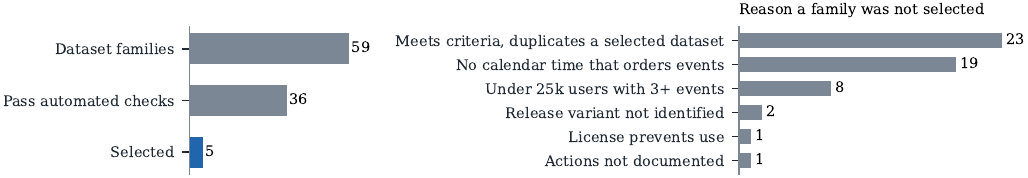}
\caption{Of 59 dataset families, 36 pass automated checks of ordering,
population, and actions, and 5 are selected. The largest group of exclusions
meets every criterion but duplicates a condition already covered. Ordering is
the most frequent defect: 19 families have no calendar time that orders
events.}
\label{fig:funnel}
\end{figure*}

Scale did not limit the selection (Figure~\ref{fig:funnel}). The two largest
candidates, with 4.79B and 27B events, were set aside for reasons unrelated to
size, and the retained datasets lie in the middle of the audited range
(Figure~\ref{fig:scale}). Ordering was the binding
constraint. Nineteen families fail it, among them grocery baskets indexed by
order number and click logs ordered only within a slate. It also excludes the
largest public bank dataset, 950M transactions from over 2M corporate clients
\cite{mollaev2025mbd}, whose release adds random noise to dates for
anonymization, which its authors note may reorder a client's events. The other
23 exclusions meet every criterion but repeat a condition a selected dataset
already covers, such as a second Amazon category; of two such datasets we keep
the one with longer per-user histories.

\begin{table}[t]
\caption{The five selected datasets. Events are raw counts. Horizon is the
number of days from the test cutoff to the last event.}
\label{tab:datasets}
\footnotesize
\setlength{\tabcolsep}{2.5pt}
\begin{tabular}{@{}lllrr@{}}
\toprule
Dataset & Condition it adds & Cutoff & Horizon & Events \\
\midrule
Amazon H\&K & sparse, 25 years & 2022-09 & 377 & 67.4M \\
KuaiRand & 8 action types, random exposure & 2022-05-06 & 2 & 323.5M \\
MicroLens & video with image, text, audio & 2022-09 & 11 & 9.1M \\
H\&M & customer profiles, price & 2020-07 & 83 & 31.8M \\
REES46 & amount, view to cart to buy & 2020-04 & 29 & 411.7M \\
\bottomrule
\end{tabular}
\end{table}

Each selected dataset adds one condition that no other provides
(Table~\ref{tab:datasets}, Figure~\ref{fig:capabilities}): Amazon Reviews 2023
Home and Kitchen \cite{hou2024amazon}, KuaiRand-27K \cite{gao2022kuairand},
MicroLens-1M \cite{ni2023microlens}, H\&M Personalized Fashion
\cite{hm2022kaggle}, and REES46 Multi-category \cite{rees46multi}. Two
conditions correspond directly to banking, where payment foundation models
treat the amount as a signal with magnitude and join static profile state to
the event sequence \cite{ostroukhov2026pragma}. REES46 records a price on
every event and a funnel from view to cart to purchase, the closest public
analogue of a transaction stream with amounts, and H\&M joins static customer
attributes to a purchase history, as a bank joins a profile to a ledger.
KuaiRand is the only source with a randomly exposed subset and explicit
negative feedback, Amazon supplies the sparse long tail, and MicroLens the
richest item content. Each dataset also invites a different shortcut (item
popularity on Amazon, the exposure policy on KuaiRand, repeat purchase on H\&M
and REES46), so no single shortcut helps on all five.

\section{Splitting in time}\label{sec:split}

A single cutoff date for all datasets would make every test cover the same
calendar period, but the five datasets admit none. H\&M and REES46 end in
2020, and KuaiRand begins in 2022 (Figure~\ref{fig:calendar}). Of the
9{,}238 daily boundaries between 1998 and 2023, none leaves events on both
sides for all five datasets, and none puts 75\% to 85\% of the rows before it
for more than two (Figure~\ref{fig:feasibility}). A date in the overlap of
Amazon and KuaiRand, 2022-05-04, leaves H\&M and REES46 with no future events
and MicroLens with 5\% of its rows in training.

Comparability therefore rests on a shared rule, fixed before any model is
evaluated. The rule targets 80\% of all rows in training and admits 75\% to
85\%. Held-out users never enter training, so about 89\% of events must
precede the cutoff to reach 80\% of rows. Candidates are month boundaries,
which stay stable under small revisions of the data, or day boundaries for
sources spanning 62 days or fewer, and the cutoff is the midpoint of the
admissible candidates (Figure~\ref{fig:sweep}). Applied again inside the
training period, the same rule sets the validation cutoff, so every test cell
has a validation counterpart. No model score enters either choice.

A shared rule still yields different horizons: 2 days on KuaiRand and 377 days
on Amazon (Table~\ref{tab:datasets}), so a difference between the two datasets
may reflect the horizon as much as the platform. INTENT therefore reports with
every result the cutoff, the length of training histories, the share of
evaluation users and target items seen in training, and the horizon.

\begin{figure}[t]
\centering
\includegraphics[width=0.92\columnwidth]{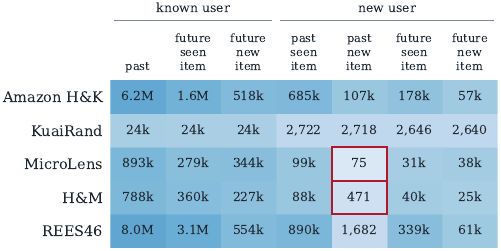}
\caption{Distinct users per dataset and test cell at each dataset's cutoff, on
a log scale. Red outlines mark the
two cells below the 1{,}000-user floor: new users choosing new items before
the cutoff on MicroLens and H\&M.}
\label{fig:census}
\end{figure}

The cutoff rule populates every cell (Figure~\ref{fig:census}). Summed over the
five datasets, the smallest test cell (new user, past, new item) holds
111{,}769 users and the largest (known user, past) 15.9M, and the validation
cells follow the same pattern. The floor applies to the sum: a dataset below
it in one cell still contributes, and its own result there is reported with
its interval but not used to rank models. Only H\&M and MicroLens fall below
the floor, both in the new-user, past, new-item cell. We retain the 10\% holdout rather than enlarge it to fill those cells,
which would reduce the population of training users.

\begin{figure*}[t]
\centering
\includegraphics[width=0.9\textwidth]{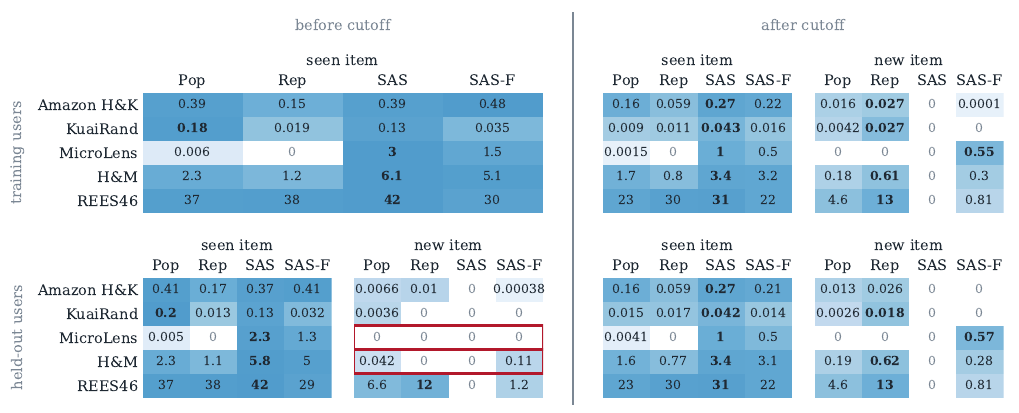}
\caption{NDCG@10 ($\times100$, mean over users) of popularity (Pop),
repeat-last (Rep), SASRec (SAS), and SASRec-F (SAS-F), laid out as in
Figure~\ref{fig:cells}. Bold marks the best model in a cell when its 95\%
confidence interval overlaps no other model's. Red outlines mark datasets
below the cell's user floor; their scores do not rank models. Shading is
log-scaled to each dataset's best score over three decades, and white is zero.
Catalogue sizes differ by a factor of 350, so scores are not comparable across
datasets.}
\label{fig:results}
\end{figure*}

\section{Reference results}\label{sec:baselines}

We evaluate four reference models. \emph{Popularity} mixes the user's own item
histogram (weight 0.7) with the smoothed global item distribution of training
events. The user's histogram starts from their history before the cutoff and
adds each later event once it has been predicted, so no prediction sees its
own target. \emph{Repeat-last} ranks the user's previous item first and ties
every other item below it. \emph{SASRec} \cite{kang2018sasrec} is a causal
transformer with two blocks, 64 dimensions, and a context of 50 events,
trained for 20{,}000 steps of 256 sequences on every dataset; its tied item
embeddings score every item seen in training. \emph{SASRec-F} (SAS-F) replaces
each item vector with the mean of learned embeddings of its metadata tokens,
such as title, category, and brand words drawn from training items, plus an
identifier embedding for a training item. Training drops the identifier
embedding with probability 0.5, so the model learns to rank items from their
metadata alone and can score a new item from its tokens. At evaluation the
identifier embedding is scaled by 0.5, its expected value in training.
All hyperparameters, including the popularity weight, are set before any test
score is computed. Each model ranks the full catalogue, predicts every target
from earlier events only, and is scored on the same targets: the first 100
events of each user per split that follow an earlier event of that user. The
known-user, past cell repeats training events, so it is scored on a fixed
random sample of 10{,}000 training users per dataset. A tied target takes the
expected rank of its tie group.

\emph{The cell determines the best model} (Figure~\ref{fig:results}). For
known users choosing seen items after the cutoff, SASRec leads on every
dataset and doubles popularity on H\&M (3.4 against 1.7). When the same users
choose new items, SASRec scores zero, and repeat-last leads on every dataset
except MicroLens, where only SASRec-F scores above zero. New users after the
cutoff show the same reversal.

\emph{Time costs more than a new user.} After the cutoff, SASRec scores within
4\% of its known-user score for new users on every dataset, and no model
differs significantly between the two. None of the models has a per-user
parameter, so a new user's history serves as well as a known user's. Crossing
the cutoff instead costs SASRec 25\% to 68\% of its score on seen items,
measured on new users so that both periods are out of sample.

\emph{Most of the time cost comes from the data.} Repeat-last fits no
parameters, so it cannot overfit the training period, yet it loses 21\% to
64\% of its score across the cutoff on Amazon, H\&M, and REES46: behavior
after the cutoff is harder to predict from history, whatever the model. A
model's loss beyond repeat-last's is therefore its own. SASRec keeps 73\% of
its score on Amazon, against 36\% for repeat-last, but 59\% on H\&M, against
71\%. On KuaiRand, popularity keeps 8\% while repeat-last does not change
significantly, which suggests that the videos popular in training are not the
ones watched after the cutoff.

\emph{Item metadata helps mainly on MicroLens.} SASRec scores zero on items
absent from training, so its memorization gap equals its whole score.
For known users after the cutoff, SASRec-F retains 112\% of its seen-item
score on MicroLens, but 10\% on H\&M and 4\% on REES46; new users show
the same pattern. It scores near zero on Amazon despite product titles, and
zero on KuaiRand, whose coarse metadata ties new videos in large groups.
On seen items after the cutoff it scores 7\% to 66\% below SASRec. Repeat-last recovers a new item
when the user interacted with it earlier in the evaluation period, but it
still loses 20\% to 57\% on new items on Amazon, H\&M, and REES46, because a
user's first choice of a new item cannot be predicted from history.

\section{Discussion and Limitations}\label{sec:discussion}

\emph{Implications for banks.} A deployed bank model meets merchants that
opened after its training ended. Identifier-only models face the limit SASRec shows here. Metadata helps
little on four datasets, but SASRec-F leads on new items on MicroLens,
which has rich item content. A bank can measure this on
its own ledger with the same protocol: a preset cutoff rule, a customer
holdout, a count of users in each cell, and a training-free reference such as
repeat-last, which separates drift in customer behavior from a model's failure
to follow it.

\emph{Limitations.} Several conditions rest on one dataset each (the
multi-year horizon on Amazon, randomized exposure and negative feedback on
KuaiRand, the purchase funnel on REES46, customer profiles on H\&M), so a
result on any of them is confounded with its platform. REES46 scores are high
partly because its funnel repeats one product across view, cart, and purchase
events. KuaiRand logs several feedback signals per impression, and the release
keeps one per event, which discards 69\% of likes and over 90\% of follows and
comments. The reference models are untuned, and only SASRec-F uses item
content, so they mark a lower reference for each cell. None has a per-user parameter, which leaves the user
axis untested. The splits have not yet passed the full release checks of
leakage tests and checksums. Three datasets forbid redistribution, so INTENT
releases processing code rather than data. No dataset comes from a bank, because the public bank data we found
fails the ordering criterion (Section~\ref{sec:select}).

\emph{Future work.} Beyond MicroLens, the limited gains from metadata motivate models that
bring item knowledge from outside the training split: pretrained content encoders, or models pretrained on other platforms that
embed a new item from its features without retraining
\cite{bruelgabrielsson2026behaviorgptv4}. Training on all five mixes calendar
periods, since Amazon's training data extends past the 2020 test periods of
H\&M and REES46, so transfer needs pairs that share an admissible cutoff
(Amazon with KuaiRand or MicroLens, H\&M with REES46). Scoring more models would test whether cell scores predict downstream
performance in fraud detection, credit scoring, and personalization.
Comparing leave-last-out splits with the known-user, future cell would
measure how much the usual protocol changes scores and rankings.

\begin{acks}
\textbf{Author contributions.} R.B.-G. conceived the project, designed the
benchmark and its evaluation protocol, and led and supervised the work. M.K.,
E.L.H., and E.d.C. carried out the majority of the experiments and built the
benchmark pipeline, with D.S. supporting them; M.K. led its implementation.
A.F. carried out the SASRec-F experiments.

\textbf{Use of AI.} Claude Code, OpenAI Codex, and Cursor assisted with
implementing and modifying code across the research codebase. Multiple
underlying models were used, and tool use was distributed across development
rather than recorded separately for each component. The authors reviewed and
tested the generated code.

We thank Gabriel Melin and Alexander Junco Hagberg of Unbox AI for their
support throughout the project.
\end{acks}

\bibliographystyle{ACM-Reference-Format}
\bibliography{refs}

\appendix
\section{Additional figures}\label{app:figs}

\begin{table*}[t]
\caption{Each audited dataset family, represented by one examined release,
with its recorded counts of users having at least three events and total
events, and the selection decision. Counts precede benchmark cleaning and
splitting and are rounded for display. D: meets every
criterion but duplicates a condition a selected dataset covers. O: no calendar
time that orders events. P: under 25{,}000 users with three or more events. V:
release variant not identified. L: license prevents use. A: actions not
documented.}
\label{tab:families}
\centering
\scriptsize
\begin{tabular}[t]{@{}lrrl@{}}
\toprule
Family & Users 3+ & Events & Decision \\
\midrule
Amazon Home \& Kitchen & 7.1M & 67M & selected \\
H\&M Fashion & 1.1M & 32M & selected \\
KuaiRand-27K & 27k & 323M & selected \\
MicroLens 1M & 1.0M & 9.1M & selected \\
REES46 Multi & 10M & 412M & selected \\
Adressa & 246k & 3.1M & D \\
Avazu & 571k & 40M & D \\
Behance & 39k & 1.0M & D \\
Brightkite & 39k & 4.7M & D \\
Diginetica & 162k & 1.2M & D \\
Food.com & 38k & 1.1M & D \\
Globo & 221k & 3.0M & D \\
Goodreads & 834k & 229M & D \\
Google Local 2021 & 37M & 662M & D \\
Gowalla & 94k & 6.4M & D \\
KuaiSAR & 26k & 11M & D \\
LibraryThing & 37k & 1.1M & D \\
MIND & 48k & 5.8M & D \\
MovieLens 25M & 163k & 25M & D \\
Music4All-Onion & 72k & 122M & D \\
Netflix Prize & 477k & 100M & D \\
Otto & 12M & 231M & D \\
RecSys Challenge 2015 & 8.0M & 66M & D \\
REES46 Cosmetics & 630k & 21M & D \\
Retailrocket & 200k & 2.8M & D \\
Steam Reviews & 689k & 7.8M & D \\
Taobao & 988k & 100M & D \\
Twitch & 100k & 3.1M & D \\
Alibaba iFashion & 3.2M & 19M & O \\
Anime & 70k & 7.8M & O \\
\bottomrule
\end{tabular}
\hspace{3em}
\begin{tabular}[t]{@{}lrrl@{}}
\toprule
Family & Users 3+ & Events & Decision \\
\midrule
Book Crossing & 14k & 494k & O \\
Booking.com & 200k & 1.2M & O \\
CiteULike & 7.9k & 135k & O \\
Criteo Attribution & 1.8M & 16M & O \\
Ecommerce User Behavior & 9.6k & 288k & O \\
Instacart & 206k & 34M & O \\
iPinYou & 559k & 19M & O \\
Jester & 25k & 1.8M & O \\
KDD Cup 2010 & 9.3k & 29M & O \\
KGRec (sound) & 20k & 2.1M & O \\
Last.fm (HetRec) & 1.9k & 93k & O \\
MOOC & 7.0k & 412k & O \\
Pinterest & 55k & 1.4M & O \\
RecGaze & 87 & 3.2k & O \\
RecSys Slates & 2.2M & 28M & O \\
Tenrec & 6.1M & 542M & O \\
Wikipedia Clickstream & 2.2M & 7.1B & O \\
BeerAdvocate & 19k & 1.6M & P \\
Douban & 2.7k & 1.7M & P \\
Epinions & 22k & 509k & P \\
Foursquare & 2.3k & 801k & P \\
KuaiRec & 7.2k & 13M & P \\
ModCloth & 7.4k & 83k & P \\
NineRec & 6.8k & 140k & P \\
Rent the Runway & 16k & 193k & P \\
PixelRec & 14M & 196M & V \\
Ta Feng & 29k & 818k & V \\
Yelp & 530k & 7.0M & L \\
Amazon Q\&A & 28k & 1.3M & A \\
\bottomrule
\end{tabular}

\end{table*}

\begin{figure}[H]
\centering
\includegraphics[width=\columnwidth]{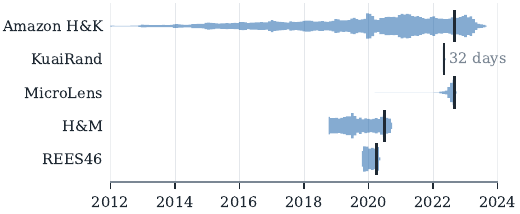}
\caption{Monthly event density of the five datasets, with each test cutoff
marked by a black bar. Amazon begins in 1998; the axis starts in 2012. No date
lies within all five spans.}
\label{fig:calendar}
\end{figure}

\begin{figure}[H]
\centering
\includegraphics[width=\columnwidth]{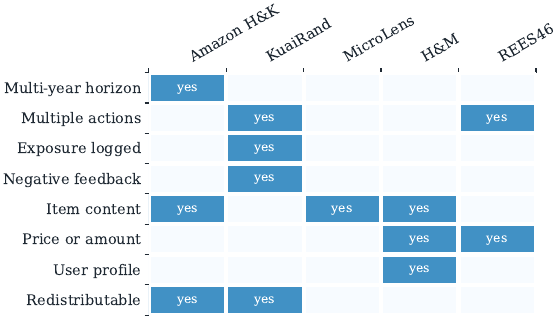}
\caption{Conditions each selected dataset provides. No two rows are equal, and
four conditions (multi-year horizon, logged exposure, negative feedback, user
profile) rest on a single dataset. Three of the five datasets cannot be
redistributed.}
\label{fig:capabilities}
\end{figure}

\begin{figure}[H]
\centering
\includegraphics[width=\columnwidth]{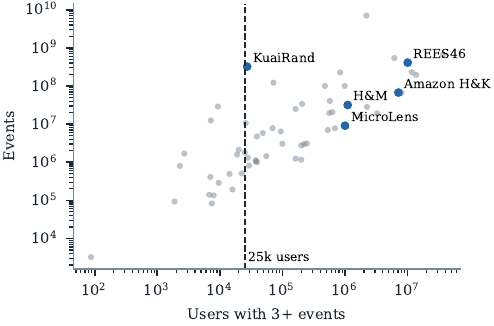}
\caption{Users with three or more events against total events for every
dataset family with both counts. The five selected datasets (blue) lie in the
middle of the range, and KuaiRand sits just above the 25{,}000-user threshold.
Many larger families were excluded for ordering or duplication.}
\label{fig:scale}
\end{figure}

\begin{figure}[H]
\centering
\includegraphics[width=\columnwidth]{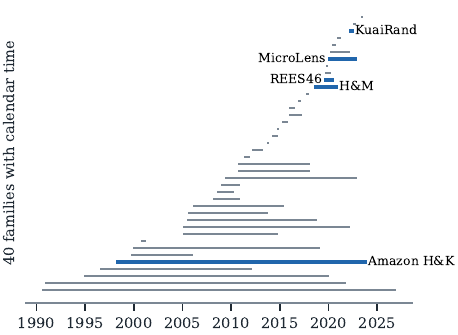}
\caption{Calendar span of the 40 families that carry calendar time, ordered by
start date. Public behavioral data spans three decades, but no single period
covers the five selected datasets (blue).}
\label{fig:coverage}
\end{figure}

\begin{figure}[H]
\centering
\includegraphics[width=\columnwidth]{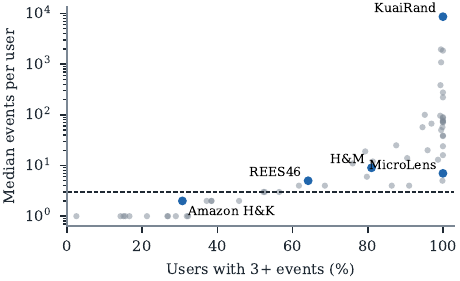}
\caption{Median events per user against the share of users with three or more
events, for every family. The dashed line marks three events. Eleven families
have a median of one event per user, and in 16 the three-event rule keeps
fewer than half the users, so the criterion counts retained users rather than
raw activity.}
\label{fig:depth}
\end{figure}

\begin{figure}[H]
\centering
\includegraphics[width=\columnwidth]{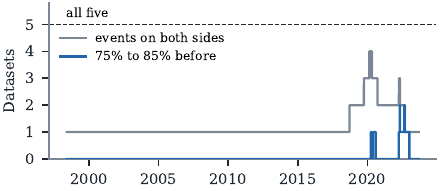}
\caption{Datasets that a single shared cutoff could serve, at each of the
9{,}238 daily boundaries from 1998 to 2023. At most four datasets have events
on both sides of any date, and at most two have 75\% to 85\% of their rows
before it. No date serves all five.}
\label{fig:feasibility}
\end{figure}

\begin{figure}[H]
\centering
\includegraphics[width=\columnwidth]{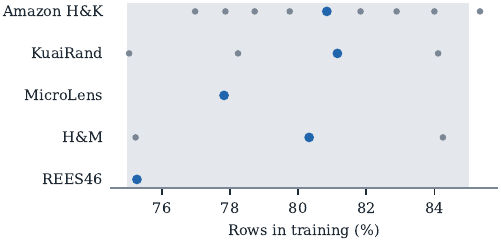}
\caption{Share of rows in training after cleaning at every admissible cutoff,
with the selected cutoff in blue and the 75\% to 85\% target band shaded. The
rule admits candidates from raw daily counts, so a measured share can fall just
outside the band. MicroLens and REES46 each admit a single month boundary.}
\label{fig:sweep}
\end{figure}

\end{document}